\documentclass{article} 
\usepackage{iclr2027_conference,times}

\usepackage{amsmath,amsfonts,bm}

\def\eqref#1{equation~\ref{#1}}

\def\1{\bm{1}}

\DeclareMathAlphabet{\mathsfit}{\encodingdefault}{\sfdefault}{m}{sl}
\SetMathAlphabet{\mathsfit}{bold}{\encodingdefault}{\sfdefault}{bx}{n}

\usepackage{amsthm}
\usepackage{amssymb}
\usepackage{amsmath}
\usepackage{algorithm}
\usepackage{enumitem}
\usepackage{picinpar}
\usepackage{lineno}
\usepackage{graphicx}
\usepackage[table]{xcolor}
\usepackage{algorithmicx}
\usepackage[noend]{algpseudocode}
\usepackage{subfigure}
\usepackage{multirow}
\usepackage{color}
\usepackage{balance}
\usepackage{enumitem}
\usepackage{hhline}
\usepackage[normalem]{ulem}
\usepackage{booktabs}
\usepackage{wrapfig}
\usepackage{cancel}
\usepackage{hyperref}
\usepackage{makecell}
\usepackage{hyperref}
\usepackage{url}

\newcommand{\stitle}[1]{\vspace{1mm} \noindent {\bf #1}}
\newcommand{\model}{ACE-Trader}
\providecommand{\best}[1]{\textbf{#1}}
\providecommand{\second}[1]{\underline{#1}}

\title{ACE-Trader: Adversarial Co-Evolution for Generator–Critic Trading Systems}

\iclrfinalcopy 

\author{
\textbf{Chang Zhou}$^{1}$ \
\textbf{Jiarun Zhou}$^{1}$ \
\textbf{Xingtong Yu}$^{2}$\thanks{Corresponding authors.} \
\textbf{Hong Cheng}$^{2}$ \
\textbf{Hui Zhang}$^{1}$ \
\textbf{Xinming Zhang}$^{1}$\footnotemark[1] \\[0.5em]
$^{1}$University of Science and Technology of China \\
$^{2}$The Chinese University of Hong Kong
}

\begin{document}

\maketitle
\pagestyle{plain}
\thispagestyle{plain}

\begin{abstract}
Generator--critic trading systems pair an LLM-based trader that makes trading decisions with a red team agent that evaluates these decisions and provides feedback for revision. However, improving individual trading decisions does not ensure that both agents learn from experience and improve together over time. This raises an important question: how can the trader and red-Team select useful experience and learn from each other's evolving behavior? On one hand, experience that benefits decisions under one market condition may become ineffective or harmful under another. On the other hand, trading outcomes reflect both agents' actions, making it difficult to derive appropriate experience for each role. In this work, we propose ACE-Trader, an adversarial co-evolution trading framework. Specifically, we first develop separate context-dependent memory selectors that learn to identify useful experience for the current trading decision. Both selectors receive shared feedback from net-return differences between complete memory-assisted and memory-free trading pipelines, with online updates emphasizing recent outcomes. We further design a co-evolution mechanism that links role-specific memories through shared trading records. We use these records to evaluate the Red-Team's risk assessments and the effects of the Trader's revisions against realized market outcomes, then derive experience for their respective memories. By linking this experience to the corresponding records, we enable the Trader to improve its proposals based on prior critiques and the Red-Team to adjust its review focus as the Trader's behavior changes. Experiments on U.S. stock trading show that ACE-Trader outperforms all 22 active baselines in cumulative return and achieves the highest Sharpe ratio and lowest maximum drawdown across both active methods and passive benchmarks.
\end{abstract}

\section{Introduction}\label{sec.intro}
\begin{figure}[t]
    \centering
    \includegraphics[width=\linewidth]{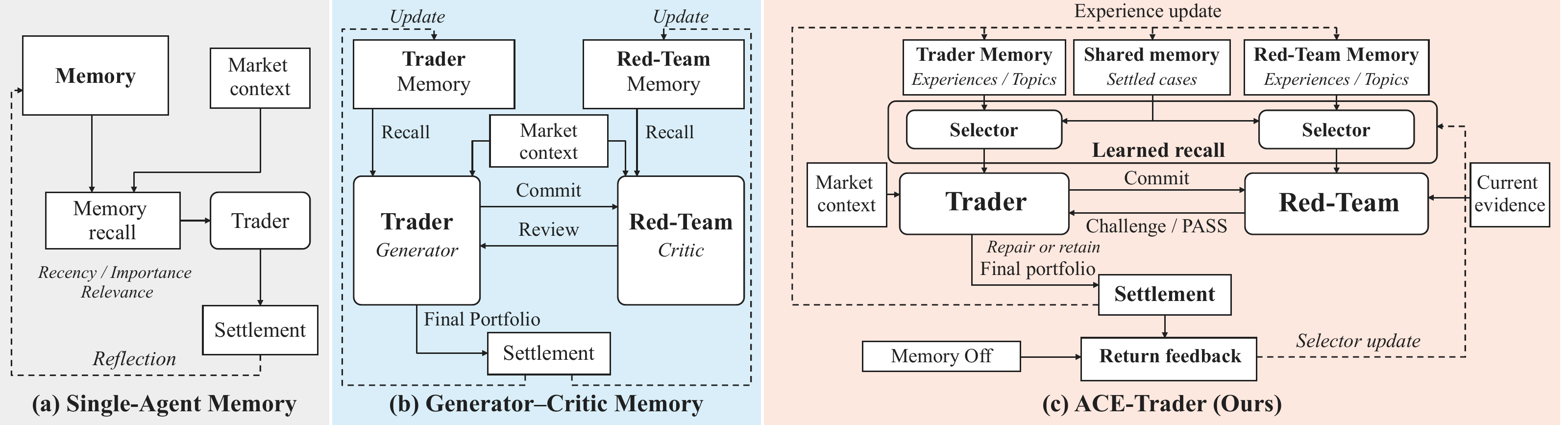}
    \caption{\textbf{Comparison of memory mechanisms in trading agents.}}
    \label{fig:motivation}
\end{figure}

Recent advances in large language models (LLMs) \citep{brown2020language,ouyang2022training} and agentic systems \citep{yao2023react,wu2024autogen} have spurred growing interest in building autonomous agents for financial trading. 
Existing LLM-based trading agents analyze market information and make trading decisions based on their analyses \citep{yu2025finmem,yu2024fincon,zhang2024finagent,xiao2024tradingagents}.
To improve subsequent trading decisions, reflection-based trading agent \citep{yu2025finmem,zhang2024finagent} reviews past decisions and their realized outcomes to identify useful lessons, which they store in memory to guide subsequent decisions, as shown in Fig.~\ref{fig:motivation}(a) . 
However, self-reflection may overlook weaknesses in the assumptions underlying the agent's own decisions\citep{liang2024encouraging}. Generator--critic systems address this limitation by assigning decision-making and critique to different agents \citep{gong2025unsupervised}. In financial trading, a trader agent serves as the generator, proposing trading decisions and their rationales, while a red-team agent serves as the critic, evaluating these proposals and providing feedback for revision before execution \citep{xiao2024tradingagents}, as shown in Fig.~\ref{fig:motivation}(b).
Improving this collaboration over time requires both agents to learn from their trading interactions and use the resulting experience to guide subsequent decisions.

However, reflecting on the overall trading outcome can lead agents to store misleading lessons in their respective memories: a profitable trade may conceal an inappropriate revision by the trader or an incorrect critique by the red-team, while a losing trade may obscure a useful red-team warning.\citep{foerster2018counterfactual} Even when appropriate lessons are extracted, they may not benefit subsequent decisions under different market conditions\citep{ang2002international,zhao2023doubleadapt}.
Yet common memory retrieval mechanisms select experience based on recency, importance, or semantic relevance \citep{yu2025finmem,zhang2024finagent}, without directly assessing its usefulness for the current trading decision. For example, recalling a lesson about reducing exposure during a downturn may lead an agent to remain overly cautious during a market recovery. These limitations raise a central question: \emph{how can a trader and a red-team learn appropriate lessons for their respective roles and select experience that benefits the current decision?} The problem is non-trivial due to two key challenges.

First, \emph{how can  Trader and Red-Team adapt memory selection to changing market conditions?}
The usefulness of past experience depends on the current market state. For the Trader, buying after a price decline may be useful in a range-bound market but lead to further losses during a sustained downtrend. For the Red-Team, a warning to reduce exposure during high volatility may prompt unnecessary position reductions once volatility subsides. Selecting useful experience therefore requires assessing its value in the current context, which cannot be determined from its past success alone. We propose separate lightweight, context-dependent memory selectors for the Trader and the Red-Team that learn which experience benefits the current trading decision. Each selector combines a frozen semantic encoder with an ensemble of trainable feedforward networks to score candidate memories using the current decision context, including market conditions, together with memory content and recorded outcomes. The highest-scoring candidate is selected for the corresponding agent. Both selectors receive shared feedback from the difference in net returns between the complete memory-assisted and memory-free trading pipelines. They are updated separately after trading outcomes are realized, with greater weight assigned to recent outcomes to adapt selection to changing market conditions.

Second, \emph{how can Trader and Red-Team learn from each other's evolving behavior rather than update their experience in isolation?}
While memory selection determines which past experience guides the current trading decision, the Trader and Red-Team must also learn from the proposal, critique, and subsequent revision to improve future decisions.
For example, a Red-Team warning may teach the Trader to avoid a previously overlooked concentration risk in later portfolios. Once the Trader corrects this behavior, the Red-Team should adjust its review focus rather than repeat warnings about a problem that has already been addressed. Such adaptation requires each agent to learn not only from market outcomes but also from the other agent's actions and responses. However, reflecting separately may lead the Trader to overlook how the Red-Team's warning shaped its revision, and the Red-Team to judge its warning without considering how the Trader responded.
In this work, we propose a memory co-evolution mechanism that links the Trader's and Red-Team's role-specific experience through shared memory, enabling each agent to adapt to the other's evolving behavior.
We design a shared case memory to record the Trader's original proposal, the Red-Team's critique, the Trader's response, and the realized market outcome. To guide role-specific memory updates, we use these records to evaluate the Red-Team's risk assessments and the effect of the Trader's revisions on portfolio returns. Based on these evaluations, we derive experience about portfolio construction and revision for the Trader and risk assessment for the Red-Team. We store this experience in their respective memories and link it to the corresponding shared memory, so each agent can revisit how the other acted and responded. We further consolidate recurring experience into long-term memory with explicit applicability conditions. When selected for subsequent decisions, this experience helps the Trader improve its proposals in response to prior critiques and the Red-Team adjust its review focus as the Trader's behavior changes.

In summary, our contributions are fourfold.
First,  we propose \textbf{ACE-Trader}, a framework for \textbf{A}dversarial \textbf{C}o-\textbf{E}volution. To the best of our knowledge, this is the first framework to explicitly support co-evolution between a Trader and a Red-Team through shared memory adaptation for financial trading.
Second, we develop lightweight, context-dependent memory selectors trained with shared net-return feedback from complete memory-assisted and memory-free trading pipelines.
Third, we design a co-evolution mechanism that links the Trader's and Red-Team's role-specific memories through shared trading cases, enabling each agent to learn from and adapt to the other's evolving behavior.
Fourth, we evaluate ACE-Trader on U.S. stock trading and use ablation studies, hyperparameter analysis, and case studies to examine its performance and the roles of its key components.

\section{Related Work}\label{sec.related}

\stitle{AI methods for trading.}
Traditional machine learning methods, including deep learning, use numerical market data to forecast returns and construct portfolios \citep{heaton2017deep,gu2020empirical,ozbayoglu2020deep}. LLM-based single-agent systems combine financial tools and language-based reasoning to analyze market information and generate trading decisions \citep{yu2025finmem,zhang2024finagent}. Generator–critic systems separate proposal generation from evaluation, using feedback from a critic to guide revisions\citep{gong2025unsupervised}. In financial trading, review and debate mechanisms similarly allow proposed decisions to be assessed and revised before execution \citep{yu2024fincon,xiao2024tradingagents}. However, revising individual proposals does not by itself ensure that the agents learn from past interactions and improve together over time.


\stitle{Memory-based adaptation for trading.}
 Memory-augmented trading agents retain experience from past decisions to guide later decisions \citep{yu2025finmem,zhang2024finagent}. They organize this experience into layered or diversified memories and retrieve it according to recency, importance, or relevance. Recently, some methods further reflect on observed trading outcomes to revise investment beliefs and risk preferences \citep{yu2024fincon}. However, retrieving a recent or relevant memory does not establish whether it will benefit the current decision under changing market conditions. Moreover, independent reflections can lose the connection between one agent's feedback and the other's response, making it difficult for the agents to learn from each other's evolving behavior. ACE-Trader combines context-dependent memory selection with shared trading records that connect the Trader's and Red-Team's experience.
\section{Preliminaries}\label{sec.prelim}


We consider daily portfolio trading over a universe of $N$ assets.
At decision step $t$, the input consists of market information $o_t$, including historical prices, news, and technical indicators, and account state $b_t$, including cash, existing holdings, net asset value, and portfolio exposure.
The output is a portfolio plan $a_t$ that specifies \texttt{BUY}, \texttt{SELL}, or \texttt{HOLD} decisions for selected assets, together with target portfolio weights expressed as percentages of the account's net asset value after transaction costs.
The total absolute exposure across long and short positions is limited to 100\% of net asset value.
All decisions and financial tool queries use only information available at the decision cutoff.

After executing $a_t$, we evaluate the realized net return $r_t$ over the settlement interval as the relative change in account value after transaction costs \citep{liu2021finrl}.
Transaction costs apply to the actual amounts traded when opening, adjusting, closing, or reversing positions.
Cash, holdings, and account value are carried forward without daily resets.
Each trading episode consists of one decision--execution--settlement cycle within this continuous account trajectory.
The objective is to improve cumulative net return subject to the portfolio constraints.

\section{Proposed Approach}\label{sec.method}
\begin{figure}[t]
    \centering
    \includegraphics[width=\linewidth]{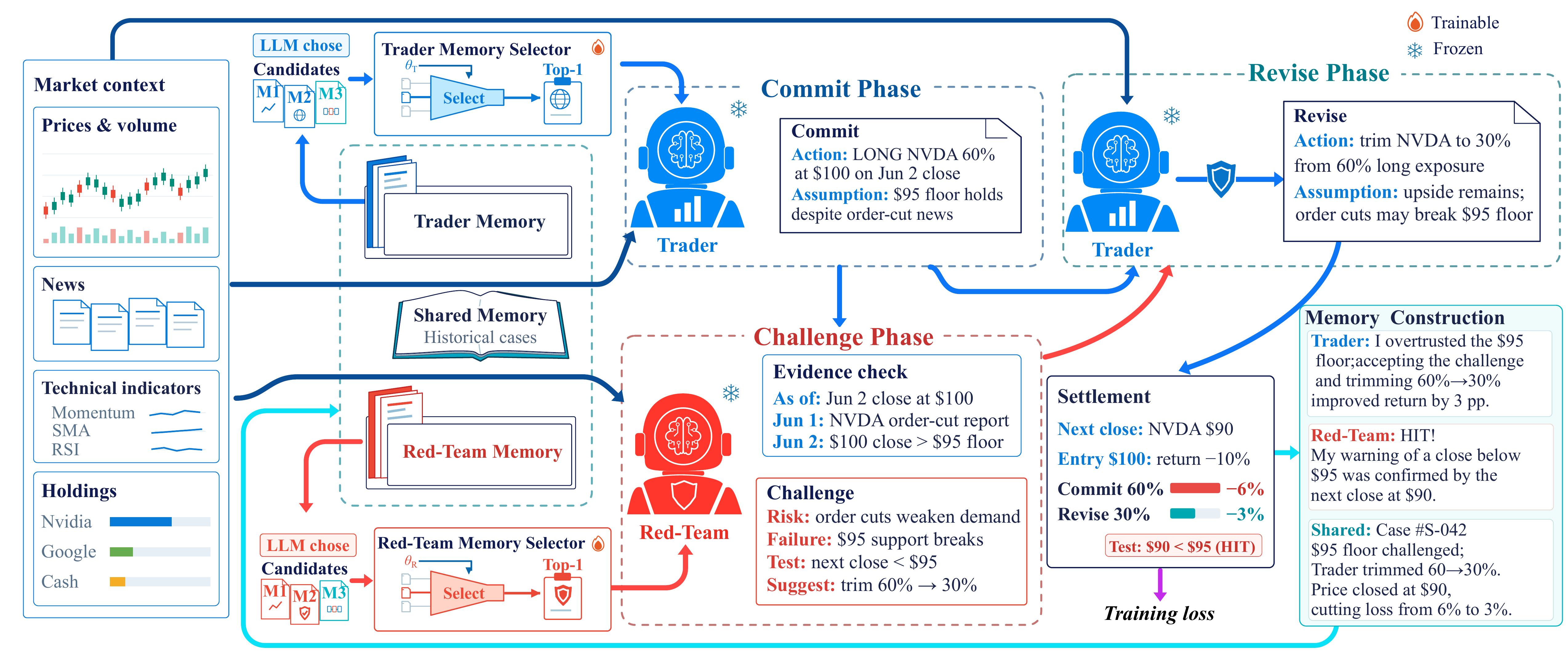}
    \caption{Overview of \model{}.}
    \label{fig:framework}
\end{figure}
\subsection{Overview}\label{sec.method.overview}
We illustrate the overall framework of \model\ in Figure~\ref{fig:framework}. 
ACE-Trader consists of two main modules: \emph{memory selection and memory-guided trading}, and \emph{memory co-evolution}.
For memory-guided trading, we use separate selectors to choose experience for the Trader and Red-Team based on the current decision context.
With memory guidance, the Trader proposes a portfolio, the Red-Team reviews the proposal using current evidence, and the Trader revises the portfolio in response.
For memory co-evolution, we use the realized outcomes to evaluate whether the Red-Team's predicted risks occurred and whether the Trader's revisions improved net return over the original proposal.
We then construct role-specific memories and shared memories which link them to shared interaction records, allowing each agent to learn from the other's actions and responses.

\subsection{Memory Selection and Memory-Guided Trading}\label{sec:memory_guided_trading}
At trading step $t$, the Trader and Red-Team receive market information $o_t$ and account state $b_t$.
Each agent maintains private memory containing its role-specific experience and can access shared memory containing past trading interactions.
We first describe how each agent selects memory, then explain how the memory guides the Commit--Challenge--Revise workflow.
\subsubsection{Selecting Useful Memories}
\stitle{Recalling candidates.}
Before the Trader commits a portfolio and before the Red-Team reviews it, each role recalls candidates for its current decision context.
We retrieve candidate memories separately for each phase $u \in \{\mathrm{Com}, \mathrm{Cha}, \mathrm{Rev}\}$, corresponding to commit, challenge, and revise phase.
Retrieval uses the current market and account information, together with any proposal, feedback, or tool evidence already available in that phase.
Candidates come from the corresponding agent's private memory and the shared memory.
We denote the resulting candidate set by $\mathcal{M}_t^u$.

\stitle{Encoding candidate representations.}
For each candidate $m\in\mathcal{M}_t^u$, we construct semantic features $\mathbf{e}_{t,m}^u$ from its text and the phase-specific retrieval context using a frozen semantic encoder $E$ and a linear projection, as detailed in Appendix~\ref{app:Selector}.
We concatenate these features with market and portfolio features and the memory's recorded outcomes:
\begin{equation}
    \mathbf{v}_{t,m}^{u}
    =
    \operatorname{Concat}
    \left(
        \mathbf{e}_{t,m}^{u},
        \mathbf{f}_{t}^{u},
        \mathbf{h}_{m}
    \right).
    \label{eq:memory_representation}
\end{equation}
Here, $\mathbf{f}_{t}^{u}$ contains current market features from $o_t$ and portfolio features specific to phase $u$.
In Commit, the portfolio features describe the actual pre-trade holdings in $b_t$.
In Challenge and Revise, they describe the portfolio that would result from applying the proposed trading actions to $b_t$, without executing those actions.
The vector $\mathbf{h}_{m}$ contains the historical outcomes recorded in memory $m$.

\stitle{Scoring and selecting memories.}
The Trader and Red-Team use separate selectors, each comprising $K$ feedforward networks.
The Trader shares its selector across Commit and Revise, but retrieves and selects memory separately in each phase.
Each network scores candidate $m$ using $\mathbf{v}_{t,m}^{u}$, and we combine the outputs as
\begin{equation}
    s_{t,m}^{u}
    =
    \mu_{t,m}^{u}
    + 
    \alpha \sigma_{t,m}^{u},
    \label{eq:memory_selection_score}
\end{equation}
where $\mu_{t,m}^{u}$ and $\sigma_{t,m}^{u}$ are the mean and standard deviation of the network outputs, and $\alpha \geq 0$ is a hyperparameter controlling the disagreement bonus.
We rank candidates by $s_{t,m}^{u}$ and supply the highest-scoring candidate as $m_t^u$.
If the candidate set $\mathcal{M}_t^u$ is empty, no memory is supplied and $m_t^u=\varnothing$.

\subsubsection{Memory-Guided Adversarial Trading}

For each phase $u$, we construct a prompt $P_t^u$ containing the phase-specific instructions and available information.
The operators $\operatorname{Trader}$ and $\operatorname{RedTeam}$ each take a prompt as input and produce their outputs through the corresponding tool-assisted workflow.

\stitle{Trader commit.}
We construct $P_t^{\mathrm{Com}}$ from market information $o_t$, account state $b_t$, and the historical observations in memory $m_t^{\mathrm{Com}}$.
The Trader uses this prompt to produce an initial trading plan and its rationale:
\begin{equation}
    \left(a_t^{\mathrm{Com}}, d_t^{\mathrm{Com}}\right)
    =
    \operatorname{Trader}
    \left(P_t^{\mathrm{Com}}\right).
    \label{eq:trader_commit}
\end{equation}
Here, $a_t^{\mathrm{Com}}$ specifies the proposed trading actions and target portfolio weights, and $d_t^{\mathrm{Com}}$ states the supporting reasons and assumptions.
For each assumption, $d_t^{\mathrm{Com}}$ records an observable condition that would invalidate it and the time horizon for checking that condition.
Both are passed to the Red-Team.

\stitle{Evidence-based challenge.}
The Red-Team selects memory $m_t^{\mathrm{Cha}}$ based on the current proposal and market information.
We construct $P_t^{\mathrm{Cha}}$ from $o_t$, $b_t$, the proposal $a_t^{\mathrm{Com}}$, its rationale $d_t^{\mathrm{Com}}$, and memory $m_t^{\mathrm{Cha}}$, supplemented with linked case details when applicable.
The Red-Team examines the proposal's assumptions and risks using current evidence gathered through news search and market-data tools:
\begin{equation}
    c_t
    =
    \operatorname{RedTeam}
    \left(P_t^{\mathrm{Cha}}\right).
    \label{eq:redteam_challenge}
\end{equation}
Here, $c_t$ contains the review evidence and either a challenge or \texttt{PASS}.
The Red-Team issues a challenge only when current evidence contradicts a specific assumption in the Trader's proposal.
Each challenge includes suggested portfolio adjustments and a risk warning specifying an observable condition and when it should be evaluated.
When the evidence is insufficient, the Red-Team returns \texttt{PASS} with any monitoring suggestions.

\stitle{Trader revise.}
The Trader retrieves and selects memory again using the original proposal, market information $o_t$, and Red-Team feedback $c_t$.
We construct $P_t^{\mathrm{Rev}}$ from $o_t$, $b_t$, $a_t^{\mathrm{Com}}$, $d_t^{\mathrm{Com}}$, $c_t$, and the newly selected memory $m_t^{\mathrm{Rev}}$.
The Trader uses this prompt to produce its final plan and response:
\begin{equation}
    \left(a_t^{\mathrm{Rev}}, d_t^{\mathrm{Rev}}\right)
    =
    \operatorname{Trader}
    \left(P_t^{\mathrm{Rev}}\right).
    \label{eq:trader_revise}
\end{equation}
Here, $a_t^{\mathrm{Rev}}$ is the final trading plan, and $d_t^{\mathrm{Rev}}$ explains the Trader's response.
The Trader revises disputed positions or provides additional current evidence to retain them.
If the Red-Team returns \texttt{PASS}, we skip Revise without further memory retrieval, selection, or a Trader call, and set $a_t^{\mathrm{Rev}}=a_t^{\mathrm{Com}}$. The final plan is $a_t=a_t^{\mathrm{Rev}}$.

\subsection{Memory Co-Evolution}
We use each trading interaction and its observed outcomes to construct experience for both roles.
By evaluating the original proposal, the Red-Team's review, and the Trader's response together, we preserve how the agents influenced each other's decisions.
The resulting role-specific memories are linked through shared cases for use in subsequent interactions.

\stitle{Settlement and evaluation.}
We execute only the final plan $a_t^{\mathrm{Rev}}$ and carry forward the resulting cash and holdings.
Once the settlement outcomes are observed, we compare its net return with the return that the original proposal $a_t^{\mathrm{Com}}$ would have produced.
Both calculations start from the same account state $b_t$ and use the same observed prices and transaction-cost rules.
Their difference indicates whether responding to the review improved net return, rather than merely whether the final plan was profitable.

We also check the risk warnings in $c_t$ against the conditions and evaluation times specified before execution.
A warning is a \emph{hit} if the observed outcomes satisfy its stated condition under the specified evaluation rule.
It is a \emph{false alarm} if its evaluation horizon has ended and sufficient observations show that the condition was not satisfied.
Warnings that cannot yet be assessed remain \emph{unresolved}.
For the assumptions in $d_t^{\mathrm{Com}}$, we check whether their stated failure conditions are met within the corresponding horizons.
An assumption that fails without a corresponding valid Red-Team challenge is recorded as a \emph{miss}.
Risk occurrence and revision benefit are evaluated separately: a correct warning can still be followed by a revision that reduces net return.

\stitle{Constructing shared and role-specific memories.}
We store the complete interaction, risk labels, and return comparison as a shared memory case.
Using predefined rules, we then extract a short-term memory item for each role.
The Trader's item records its original and final trading decisions, its response to the review, and the net-return difference between the two plans.
When the final plan underperforms the original proposal, the item explicitly records that the revision reduced net return, even if the final plan itself was profitable.
The Red-Team's item records the challenged or missed assumptions, the evidence examined, and the corresponding risk labels.
Both items include the market context and shared case identifier, preserving the connection between the proposal, critique, response, and outcome.

Each item is added to the corresponding role's private short-term memory.
Periodically, an LLM summarizes recurring lessons from recent entries for long-term memory, specifying when each lesson applies and retaining references to its supporting cases.
These lessons distinguish whether a risk occurred from whether the response improved net return.
When a lesson is recalled, the agent can revisit the linked interaction to understand the other role's actions and responses without accessing its private memory.

\subsection{Memory Selector Update}
\label{sec:selector_update}

At each trading step $t$, we run the complete trading workflow with and without memory using the same market information $o_t$ and pre-decision account state $b_t$.
We simulate the execution of their final plans on copies of $b_t$ over the same settlement interval, using identical market prices and transaction-cost rules.
Once the outcomes are observed, the net-return difference provides a shared binary label, with near-tied outcomes excluded.
Each selector is trained on the saved representations of its selected memories using recency-weighted binary cross-entropy.
The trading LLMs and semantic encoder remain frozen.
Details of supervision and the training objective are provided in Appendix~\ref{app:selector_update}.

\section{Experiments}
\label{sec:experiments}

We evaluate ACE-Trader on daily U.S. stock trading to examine portfolio performance and the contribution of memory-based adaptation under a continuous-account protocol.

\subsection{Experimental Setup}
\label{sec:experimental_setup}

\stitle{Dataset.}
We evaluate ACE-Trader on a daily U.S. stock-trading benchmark comprising 33 equities across 11 sectors.
Historical prices and financial news are drawn from FNSPID \citep{dong2024fnspid}, which contains 29.7 million stock-price records and 15.7 million time-aligned news articles.
Data from 2021-01-01 to 2023-12-31 are used for task-specific training of baselines that require it and for warming up ACE-Trader, without a separate validation set.
Testing covers 2026-01-01 to 2026-06-30, using prices, news, and associated asset information collected from Financial Modeling Prep (FMP) for the same stock universe.
The test period also falls after the documented training-data cutoffs of all evaluated LLM checkpoints.
Data construction and temporal filtering are detailed in Appendix~\ref{app:data},~\ref{app:input}.

\stitle{Baselines.}
We compare ACE-Trader with 22 active baselines organized into six categories.
(1) \emph{Machine Learning methods (ML)}: LightGBM \citep{ke2017lightgbm}, XGBoost \citep{chen2016xgboost}, LSTM \citep{hochreiter1997long}, GRU \citep{cho2014learning,chung2014empirical}, MLP \citep{hornik1989multilayer}, TRA \citep{lin2021learning}, and HIST \citep{xu2021hist}, which predict asset returns using structured market features, temporal representations, or cross-asset relationships.
(2) \emph{Reinforcement-learning trading methods (RL)}: FinRL-DQN, FinRL-A2C, and FinRL-PPO \citep{liu2021finrl}, which learn trading policies through interactions with a portfolio environment.
(3) \emph{Frontier general-purpose LLMs}: GLM 5.3 Flash, DeepSeek V4 Flash \citep{xu2026deepseek}, and Qwen3.6-35B-A3B, which directly generate trading decisions without task-specific fine-tuning.
(4) \emph{Finance-specific LLMs}: FinGPT \citep{yang2023fingpt}, FinCast \citep{zhu2025fincast}, and Kronos \citep{shi2026kronos}, which use financial-domain pretraining or adaptation to generate decisions or market forecasts.
(5) \emph{Single-agent trading systems}: FinAgent \citep{zhang2024finagent} and FLAG-Trader \citep{xiong2025flag}, which organize market analysis and decision making around one trading agent, supported by tools, memory, reflection, or policy optimization.
(6) \emph{Multi-agent trading systems}: TradingAgents \citep{xiao2024tradingagents}, LLMoE \citep{liu2025llmoe}, TradExpert \citep{ding2024tradexpert}, and MM-DREX \citep{chen2025mmdrex}, which combine specialized assessments through collaboration, debate, or expert coordination.

We also report two passive references.
\emph{Market Buy-and-Hold} follows the Nasdaq Composite throughout the evaluation period.
\emph{Equal-Weight Buy-and-Hold} allocates capital equally across the candidate stocks at the beginning of the period and holds the resulting portfolio without active rebalancing.
Further baseline descriptions are provided in Appendix~\ref{app:baselines}.

\stitle{Evaluation protocol.}
All methods are evaluated chronologically using the same information cutoffs and continuous-account simulator.
Trading decisions are made and executed at each trading day's opening, with a one-way transaction cost of 5 basis points on traded notional and a gross-exposure limit of 100\% of net asset value.
Cash and holdings persist across trading days.
ACE-Trader produces \texttt{BUY}, \texttt{SELL}, or \texttt{HOLD} decisions with target weights; baseline outputs follow the conversions in Appendix~\ref{app:execution}.
Offline training uses only historical data, without test-period hyperparameter selection.
Trading feedback is evaluated through the next common trading day's close and is used for memory and selector updates only after becoming observable.
Portfolio-return metrics use the continuous account's daily net asset values after costs.
Execution and feedback details are provided in Appendices~\ref{app:protocol} and~\ref{app:selector_update}.

\providecommand{\best}[1]{\textbf{#1}}
\providecommand{\second}[1]{\underline{#1}}

\begin{table*}[!t]
\centering
\caption{Portfolio performance on Stock.}
\label{tab:stock_main}
\resizebox{\textwidth}{!}{
\begin{tabular}{llrrrrrrr}
\toprule
Category
& Model
& Cum. Ret. (\%)$\uparrow$
& Ann. Ret. (\%)$\uparrow$
& Sharpe$\uparrow$
& MDD. (\%)$\downarrow$
& Vol. (\%)$\downarrow$
& HR. (\%)$\uparrow$
& Turnover \\
\midrule

\multirow{2}{*}{Passive}
& NASDAQ Buy \& Hold
& \best{+12.82}
& \best{+28.29}
& 1.340
& 12.84
& 20.10
& \best{57.38}
& 0.000 \\
& Equal-Weight Buy \& Hold
& +8.68
& +18.77
& \second{1.375}
& \second{5.33}
& 13.14
& 52.48
& 0.000 \\
\midrule

\multirow[t]{7}{*}{ML}
& LightGBM
& +4.86 & +10.30 & 0.502 & 13.13 & 26.48 & 50.74 & 1.341 \\
& XGBoost
& -21.18 & -38.84 & -1.610 & 29.53 & 28.07 & 50.16 & 1.329 \\
& LSTM
& +2.81 & +5.90 & 0.399 & 9.25 & 18.60 & 47.52 & 1.044 \\
& GRU
& +0.53 & +1.09 & 0.148 & 11.16 & 17.42 & 49.59 & 1.028 \\
& MLP
& -1.64 & -3.36 & -0.124 & 10.74 & 16.56 & 51.07 & 1.046 \\
& TRA
& +6.47 & +13.83 & 0.662 & 16.23 & 23.83 & 51.31 & 0.643 \\
& HIST
& +5.29 & +11.24 & 0.619 & 8.78 & 20.65 & 51.48 & 0.895 \\
\midrule

\multirow[t]{3}{*}{RL}
& FinRL-DQN
& +9.59 & +20.82 & 1.067 & 15.11 & 19.50 & 49.34 & 0.821 \\
& FinRL-A2C
& -16.84 & -31.67 & -2.245 & 18.88 & 16.35 & 47.45 & 0.915 \\
& FinRL-PPO
& -7.40 & -14.68 & -0.820 & 12.38 & 17.51 & 49.10 & 1.074 \\
\midrule

\multirow[t]{3}{*}{General LLMs}
& GLM 5.3 Flash
& -6.30 & -12.58 & -0.453 & 13.26 & 23.60 & 48.60 & 1.362 \\
& DeepSeek V4 Flash
& -19.72 & -36.47 & -1.387 & 25.05 & 29.57 & 50.41 & 1.287 \\
& Qwen3.6-35B-A3B
& -13.18 & -25.32 & -1.046 & 16.81 & 24.96 & 50.33 & 1.393 \\
\midrule

\multirow[t]{3}{*}{Financial LLMs}
& FinGPT
& -10.84 & -21.10 & -1.003 & 18.45 & 21.38 & 49.26 & 1.514 \\
& FinCast
& -1.30 & -2.66 & 0.040 & 14.68 & 27.72 & \second{54.10} & 1.636 \\
& Kronos
& -30.45 & -52.76 & -3.753 & 32.65 & 19.45 & 46.55 & 1.936 \\
\midrule

\multirow[t]{2}{*}{Single-agent}
& FinAgent
& -15.95 & -30.16 & -1.495 & 24.29 & 22.34 & 49.43 & 1.316 \\
& FLAG-Trader
& -2.62 & -5.33 & -0.159 & 15.93 & 20.86 & 50.00 & 1.370 \\
\midrule

\multirow[t]{4}{*}{Multi-agents}
& TradingAgents
& -9.42 & -18.49 & -0.769 & 19.88 & 23.11 & 50.41 & 1.343 \\
& LLMoE
& -7.33 & -14.55 & -0.795 & 13.22 & 17.80 & 47.95 & 1.337 \\
& TradExpert
& +0.10
& +0.20
& 0.072
& 6.55
& \best{10.70}
& 48.43
& 0.610 \\
& MM-DREX
& -5.27 & -10.58 & -0.615 & 13.59 & 16.10 & 48.11 & 1.350 \\
\midrule

Ours
& \best{ACE-Trader}
& \second{+12.44}
& \second{+27.40}
& \best{2.023}
& \best{5.21}
& \second{12.35}
& \best{57.38}
& 0.153 \\
\bottomrule
\end{tabular}
}
\vspace{1mm}
\begin{minipage}{\textwidth}
\footnotesize
The best and second-best results across all methods, including passive
benchmarks, are shown in \best{bold} and \second{underlined}, respectively.
Turnover is reported without ranking.
\end{minipage}
\end{table*}

\stitle{Evaluation metrics.}
We report cumulative return \citep{markowitz1952portfolio}, annualized return \citep{markowitz1952portfolio}, Sharpe ratio \citep{sharpe1966mutual}, maximum drawdown(MDD) \citep{magdon2004maximum}, annualized volatility \citep{markowitz1952portfolio}, hit rate(HR) \citep{pesaran1992simple}, and average daily turnover \citep{demiguel2009optimal}
All portfolio-return metrics are calculated from the continuous account trajectory after transaction costs.
Their definitions are given in Appendix~\ref{app:metrics}.

\stitle{Implementation details.}
Both the Trader and Red-Team use DeepSeek V4 Flash, with MiniLM encoding candidate memories.
Each selector comprises $K=3$ feedforward networks, and we set $\alpha=\gamma=0.8$.
The trading LLMs and semantic encoder remain frozen.
Further implementation details for ACE-Trader and the baselines are provided in Appendix~\ref{app:implementation}.

\subsection{Performance Evaluation}
\label{sec:performance_evaluation}
Table~\ref{tab:stock_main} reports the portfolio performance on Stock.
We make three observations.
First, ACE-Trader achieves the highest cumulative return among active methods, outperforming the strongest baseline, FinRL-DQN, by 2.85 percentage points.
This result supports the effectiveness of the complete framework for portfolio trading.
Second, ACE-Trader outperforms direct prompting with the same DeepSeek V4 Flash backbone, increasing cumulative return from -19.72\% to 12.44\%.
This comparison supports the value of the memory-guided Trader--Red-Team workflow beyond using the backbone alone.
Third, ACE-Trader achieves the highest Sharpe ratio and lowest maximum drawdown across all methods, including passive benchmarks.
Although its cumulative return is slightly below NASDAQ Buy-and-Hold, its lower drawdown and volatility indicate a more favorable risk--return trade-off.

\subsection{Ablation Study}
\begin{table}[t]
\centering
\caption{Ablation study of memory selection and co-evolution.}
\label{tab:ablation}
\resizebox{\linewidth}{!}{
\begin{tabular}{lcccrrrrr}
\toprule
Model
& TS
& RS
& CE
& Cum. Ret. (\%)$\uparrow$
& Sharpe$\uparrow$
& MDD. (\%)$\downarrow$
& Vol. (\%)$\downarrow$
& HR. (\%)$\uparrow$
 \\
\midrule

Variant 1
& $\times$
& $\times$
& $\times$
& +5.53 & 0.823 & 7.81 & 14.84 & 53.38  \\

Variant 2
& $\checkmark$
& $\times$
& $\times$
& +6.90 & 1.028 & 7.52 & 14.41 & 58.20 \\

Variant 3
& $\times$
& $\checkmark$
& $\times$
& +8.56 & 1.261 & 6.33 & 14.25 & 62.30 \\

Variant 4
& $\checkmark$
& $\checkmark$
& $\times$
& +10.07 & 1.529 & 6.36 & 13.57 & 55.74 \\

ACE-Trader
& $\checkmark$
& $\checkmark$
& $\checkmark$
& +12.44 & 2.023 & 5.21 & 12.35 & 57.38 \\

\bottomrule
\end{tabular}
}
\begin{minipage}{\linewidth}
\footnotesize
TS: Trader selector; RS: Red-Team selector;
CE: Co-Evolution.
$\checkmark$ and $\times$ indicate whether a component is included
or excluded, respectively.
\end{minipage}
\end{table}
To assess the contributions of context-dependent memory selection and memory co-evolution, we compare ACE-Trader with four variants in Table~\ref{tab:ablation}.
All variants retain the Commit--Challenge--Revise workflow.
Variants 1--4 disable memory co-evolution and use neither selector, only the Trader selector, only the Red-Team selector, or both selectors, respectively.
The complete model additionally enables memory co-evolution.
The remaining experimental settings are unchanged, and the implementation of each ablated variant is detailed in Appendix~\ref{app:ablation_settings}.

We make three observations.
First, the Trader selector improves profitability even without memory co-evolution.
Variant 2 outperforms Variant 1 across all reported metrics, while Variant 4 achieves higher cumulative return and Sharpe ratio than Variant 3.
These comparisons support the value of selecting relevant experience for portfolio construction and revision.

Second, the Red-Team selector improves risk-adjusted performance.
Variant 3 outperforms Variant 1 across all reported metrics.
Adding the Red-Team selector alongside the Trader selector also improves return and Sharpe ratio while reducing maximum drawdown and volatility, as shown by Variant 4 versus Variant 2.
These results support the value of selecting relevant experience for the review process.

Third, enabling memory co-evolution improves all reported metrics relative to Variant 4, increasing cumulative return by 2.37 percentage points.
This supports the contribution of learning from linked trading experience beyond memory selection alone.
The complete model achieves the highest return and Sharpe ratio with the lowest maximum drawdown and volatility, although Variant 3 retains the highest hit rate.
\subsection{Hyperparameter Analysis}
\label{sec:hyperparameter_analysis}
\begin{wrapfigure}{r}{0.4\textwidth}
    \centering
    \includegraphics[width=\linewidth]{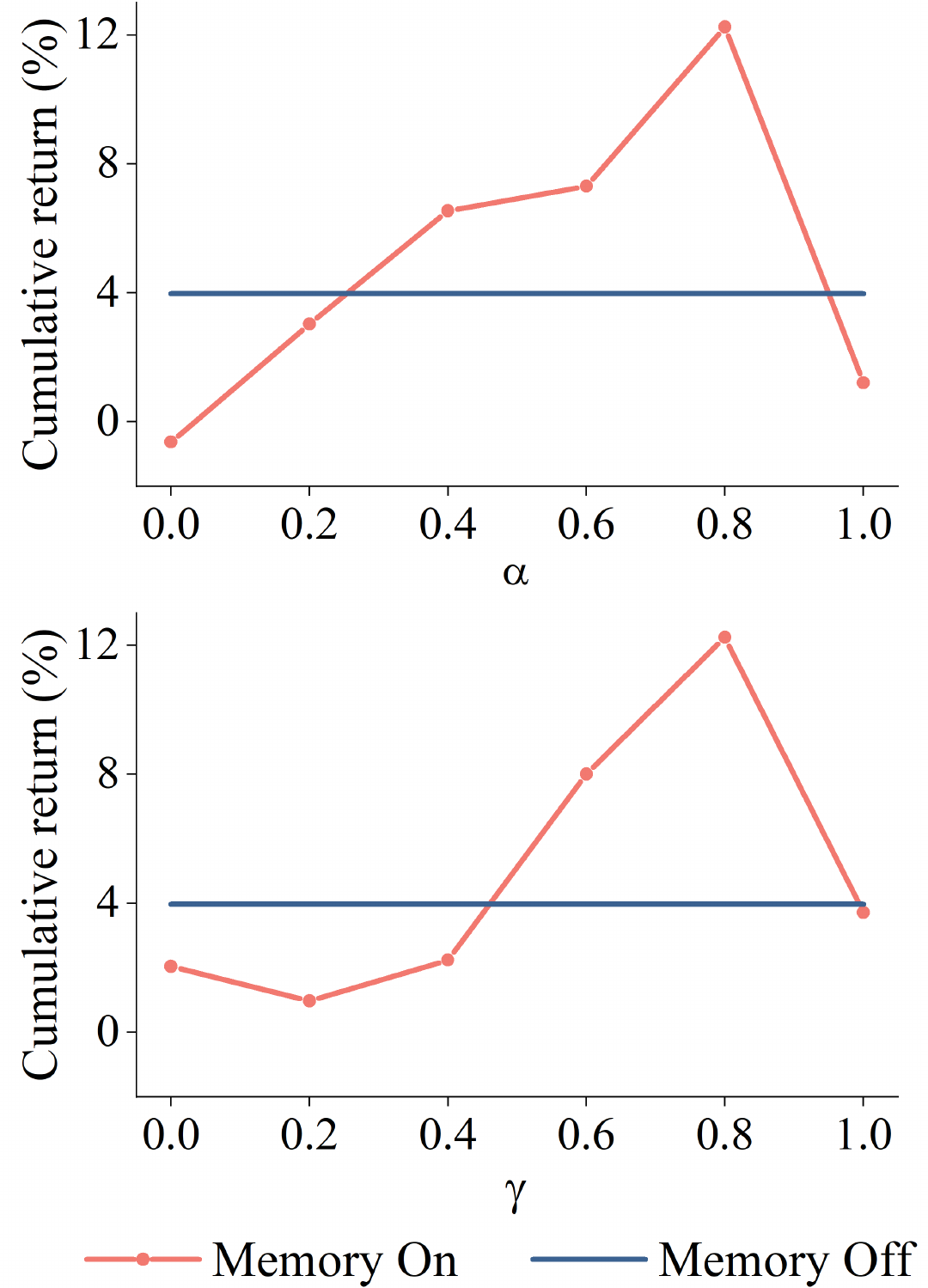}
    \caption{Impact of $\alpha$ and $\gamma$.}
    \label{fig:hyperparameters}
\end{wrapfigure}
We examine the effects of $\alpha$ in memory scoring and $\gamma$ in selector training.
When varying $\alpha$, we fix $\gamma=0.8$; when varying $\gamma$, we fix $\alpha=0.8$.
Figure~\ref{fig:hyperparameters} compares ACE-Trader (Memory On) with a memory-free baseline (Memory Off).
The baseline runs the same Commit--Challenge--Revise workflow without recalled memories and carries forward its own cash and holdings throughout testing.

\stitle{Impact of $\alpha$.}
Cumulative return increases as $\alpha$ rises from $0$ to $0.8$, but drops sharply at $1.0$. The improvement over $\alpha=0$, which ranks memories solely by their mean predictions, supports the value of disagreement-guided exploration. However, the decline at $\alpha=1.0$ suggests that placing too much weight on disagreement can be counterproductive. The highest return is achieved at $\alpha=0.8$.

\paragraph{Impact of $\gamma$.}
Small values of $\gamma$ yield limited returns. Increasing $\gamma$ gives greater weight to older feedback within the fixed training window, with the highest return achieved at $\gamma=0.8$. Performance declines at $\gamma=1.0$, where all valid trading steps within the window receive equal weight. These results suggest that using older feedback while assigning greater weight to recent outcomes is more effective than focusing only on the latest feedback or weighting all valid trading steps equally.

\subsection{Case Study: Memory Selection}
\label{sec:memory_case}

We examine the 2026-01-05 interaction, where three memory selections draw on role-specific records from the December 15, 2023 episode.
During Commit, the Trader recalls a portfolio experience from a market with broadly positive momentum to support its JPM and BHP purchases without copying historical position sizes.
During Challenge, the Red-Team selects the corresponding AVGO false-alarm record to assess whether a CAT downgrade is supported by price deterioration.
During Revise, the Trader independently reselects its portfolio record and cites the earlier false alarm in its response.
Given CAT's 2.96\% rise and positive MACD \citep{appel1985moving}, it judges the downgrade insufficient to justify reducing its existing shares.

By the January 6 close, CAT rises by 1.13\%, so the warning of a decline of at least 2.5\% is classified as a false alarm.
The original and final plans remain identical, both returning 0.15\%, with no incremental revision gain.
This case illustrates how recalled experience supports retaining a proposal when current evidence does not justify a change.
An additional case study in Appendix~\ref{app:revision_case} illustrates a beneficial portfolio revision.

\section{Conclusions}
\label{sec:conclusions}

In this paper, we proposed ACE-Trader, an adversarial co-evolution framework for generator--critic trading systems.
ACE-Trader employs separate context-dependent memory selectors to guide the Trader and Red-Team, learning from shared net-return feedback to adapt memory selection to changing market conditions.
We further developed a memory co-evolution mechanism that separately evaluates whether predicted risks occur and whether portfolio revisions improve net return, then derives role-specific experience linked through shared trading cases.
These links preserve the connection between one agent's feedback and the other's response, allowing both roles to learn from their interactions without fine-tuning the underlying LLMs.
Experiments on U.S. stock trading show that ACE-Trader outperforms
all 22 active baselines in cumulative return and achieves the highest
Sharpe ratio and lowest maximum drawdown across both active methods
and passive benchmarks.

\section*{AI Use Statement}
Generative AI tools were used to improve the readability and language of the manuscript and to assist with software code writing and editing. The research ideas, hypotheses, methodology, and experimental design were developed by the authors, and the interpretation of experimental results was conducted by the authors. All AI-assisted code and manuscript revisions were reviewed and verified by the authors.
The authors take responsibility for the final content of this work, including all AI-assisted text and code.
\section*{Reproducibility Statement}
We provide the data construction and evaluation protocol in Appendix~\ref{app:datasetandeva}, 
the settlement feedback and online update procedure in Appendix~\ref{app:selector_update}, 
and implementation and baseline configurations in Appendix~\ref{app:implementation}. 
Anonymous source code is provided with the submission to facilitate 
reproduction of the reported results.

\bibliography{iclr2027_conference}
\bibliographystyle{iclr2027_conference}

\appendix
\section{Case Study: Portfolio Revision}
\label{app:revision_case}

Table~\ref{tab:revision_case} presents the 2026-06-17 interaction, illustrating how the Red-Team's review corrects a trading rationale before execution.
The Trader treats AMZN's oversold RSI as a buying opportunity, but the Red-Team identifies a gap in this reasoning: the low RSI does not establish that the persistent downtrend is reversing.
The review confronts the rebound thesis with recent price evidence and prompts a targeted correction.
The Trader cancels only the additional AMZN purchase, retaining its existing shares and leaving the JPM and XOM sell orders unchanged.

AMZN's subsequent decline satisfies the pre-specified risk condition, confirming the warning as a hit.
In the five-trading-day fixed-holdings comparison, canceling the purchase increases net return from -0.0148\% to +0.0745\%, turning a small evaluated loss into a gain.
The improvement comes from avoiding the loss on the additional AMZN exposure and its transaction cost, rather than changing unrelated trades.
The case therefore shows how evidence-based review translates a valid risk assessment into a beneficial, targeted revision.
Together with the CAT case, it illustrates that the Trader need not automatically accept or reject challenges: it can retain a proposal when the evidence for changing it is insufficient and revise it when a specific risk is supported.

\begin{table}[t]
\centering
\caption{Case from the initial proposal to the revised decision.}
\label{tab:revision_case}
\small
\begin{tabular}{@{}p{0.20\linewidth}p{0.75\linewidth}@{}}
\toprule
Phase & Decision, evidence, and outcome \\
\midrule

\textbf{Trader: Commit}
&
\textbf{Proposal:} Buy additional AMZN worth 2\% of NAV; sell JPM worth 1\% and XOM worth 1.5\% of NAV.
\newline
\textbf{Rationale:} AMZN's RSI of 23.35 suggests an oversold rebound.
\\
\midrule

\textbf{RT: Challenge}
&
\textbf{Evidence:} AMZN's 10- and 20-day returns are -5.01\% and -8.42\%, and its price is 6.61\% below the 20-day moving average.
\newline
\textbf{Challenge:} Low RSI alone does not justify adding exposure while the downtrend persists.
\newline
\textbf{Risk condition:} AMZN's return at the end of the five-trading-day horizon is below -3\%.
\\
\midrule

\textbf{Trader: Revise}
&
\textbf{Accept the challenge and cancel the additional AMZN purchase.}
Retain the existing AMZN shares without selling them.
Keep the JPM and XOM sell orders unchanged.
\\
\midrule

\textbf{Risk evaluation}
&
AMZN returns -4.42\% over the evaluation window ending on June 25, satisfying the pre-specified risk condition.
The warning is a \textbf{hit}.
\\
\midrule

\textbf{Revision evaluation}
&
The original proposal yields -0.0148\%, whereas the revised plan yields +0.0745\%.
Canceling the purchase improves net return by \textbf{0.0893\%}, turning a small evaluated loss into a gain.
\\

\bottomrule
\end{tabular}

\vspace{1mm}
\begin{minipage}{\linewidth}
\footnotesize
\end{minipage}
\end{table}

\section{Settlement Feedback and Online Updates}
\label{app:selector_update}

\stitle{Constructing supervision.}
At each trading step $t$, we run ACE-Trader and a memory-free reference using the same market information $o_t$ and pre-decision account state $b_t$.
The reference runs its own Commit--Challenge--Revise workflow without recalling private memories or shared cases.
We shadow-execute the two final plans, meaning that each is simulated on a copy of $b_t$ over the same settlement interval using identical market prices, execution rules, and transaction-cost rules.
The reference is initialized from the current $b_t$ at every step, rather than carrying forward its previous simulated cash and holdings.
Only the memory-assisted final plan is executed in the persistent trading account.

Once the settlement outcomes are observed, we compare the net returns of the memory-assisted and memory-free plans.
We assign $\zeta_t=1$ if the memory-assisted plan outperforms the reference by more than a tolerance $\epsilon\geq 0$, and $\zeta_t=0$ if it underperforms by more than $\epsilon$.
Trading steps whose return difference lies within $[-\epsilon,\epsilon]$ are excluded from training.

Each training example pairs a selected memory's representation $\mathbf{v}_{t,m}^{u}$, saved at selection time, with $\zeta_t$.
Both selectors receive the same label but train on their own selections, including the Trader's Commit and Revise selections when both occur.
A role that selects no memory contributes no new training example for that step.
The shared label compares memory use across the complete workflow at the current account state; it does not measure the separate contribution of an individual memory.

\stitle{Recency-weighted objective.}
For role $i\in\{\mathrm{Tr},\mathrm{RT}\}$, let $\mathcal{B}_t^i$ contain the representations saved for its memory selections at trading step $t$.
A step is valid for the role when its feedback is available and not excluded, and $\mathcal{B}_t^i$ is nonempty.
The $k$-th ensemble member, parameterized by $\theta_{i,k}$, predicts the probability $p_{\theta_{i,k}}(\mathbf{v})$ of positive feedback.
Its loss for a valid trading step is
\begin{equation}
    \ell_t(\theta_{i,k})
    =
    \frac{1}{|\mathcal{B}_t^i|}
    \sum_{\mathbf{v}\in\mathcal{B}_t^i}
    \operatorname{BCE}
    \left(
        p_{\theta_{i,k}}(\mathbf{v}),
        \zeta_t
    \right),
    \label{eq:step_selector_loss}
\end{equation}
where BCE denotes binary cross-entropy.
Averaging within each trading step prevents steps with more selections from receiving extra training weight.

Each selector retains records from the $W$ most recent observed settlement days, or all observed settlement days when fewer than $W$ are available.
Only valid trading steps within this window contribute to training; days without valid feedback are not replaced by older days outside the window.
After settling step $t$, let $t_1>\cdots>t_J$ denote the contributing trading steps from newest to oldest, with $J\leq W$.
For $J>0$, the training objective is
\begin{equation}
    \mathcal{L}_t(\theta_{i,k})
    =
    \frac{
        \sum_{j=1}^{J}
        \operatorname{pow}(\gamma,d_{t,j})
        \ell_{t_j}(\theta_{i,k})
    }{
        \sum_{j=1}^{J}
        \operatorname{pow}(\gamma,d_{t,j})
    },
    \label{eq:recency_weighted_loss}
\end{equation}
where $d_{t,j}$ denotes the distance between the settlement days of steps $t_j$ and $t$, and $\operatorname{pow}(x,n)$ denotes the $n$-th power of $x$.
The weights depend on settlement-day distance rather than the rank of valid records, and the denominator normalizes their sum.
The window size $W$ and decay factor $\gamma$ are hyperparameters.

\stitle{Optimization.}
When valid feedback becomes available for trading step $t$, we update each ensemble member by minimizing $\mathcal{L}_t(\theta_{i,k})$, provided that its selector has usable records within the current window.
Feedback excluded by the tolerance rule does not trigger an update.
The saved representations and labels remain fixed, while network parameters and optimizer states are retained between updates.
The semantic encoder and trading LLMs remain frozen.
Numerical settings and implementation details are provided in Appendix~\ref{app:selector_training_settings}.

\section{Dataset and Evaluation Details}
\label{app:datasetandeva}

\subsection{Data Sources and Stock Universe}\label{app:data}

We evaluate ACE-Trader on a daily U.S. stock-trading benchmark comprising 33 equities across 11 sectors.
Historical prices and financial news are obtained from FNSPID \citep{dong2024fnspid}, a large-scale financial dataset containing 29.7 million stock-price records and 15.7 million time-aligned news articles.
The stock universe consists of U.S.-listed equities with valid OHLCV records and SEC total-asset metadata, covering large-, mid-, and small-size strata where available.

The selected tickers are
\texttt{AIT}, \texttt{AMT}, \texttt{AMZN}, \texttt{APD},
\texttt{AWR}, \texttt{BHP}, \texttt{BKNG}, \texttt{CAT},
\texttt{CMC}, \texttt{COP}, \texttt{CUBE}, \texttt{DUK},
\texttt{ELF}, \texttt{EXEL}, \texttt{FOXA}, \texttt{FSLY},
\texttt{GE}, \texttt{GOOGL}, \texttt{HOG}, \texttt{JNJ},
\texttt{JPM}, \texttt{KHC}, \texttt{MSFT}, \texttt{NRG},
\texttt{NVDA}, \texttt{RDN}, \texttt{RJF}, \texttt{RRC},
\texttt{TGNA}, \texttt{VICI}, \texttt{WMT}, \texttt{XOM},
and \texttt{ZBH}.
This asset universe remains fixed throughout training and testing.

Historical training uses data from 2021-01-01 to 2023-12-31, without a separate validation set.
Testing covers 2026-01-01 to 2026-06-30.
All data for this later period, including prices, financial news, and associated asset information, are obtained from Financial Modeling Prep (FMP).
The later test period is intended to mitigate potential temporal leakage in retrospective LLM-based trading evaluation.
Both periods use the same feature definitions, cleaning procedures, and temporal filtering rules.

\subsection{Input Construction and Causal Preprocessing} \label{app:input}

\paragraph{Prices and features.}
Price records retain raw open, high, low, close, and volume (OHLCV), together with adjusted closing prices for corporate-action auditing.
Records without a valid date or closing price are removed.
Missing open, high, or low values are replaced with the closing price from the same record, and missing trading volume is set to zero.
Prices are never interpolated across dates.

Each asset input contains its sector, market capitalization, five completed daily OHLCV bars, and technical indicators computed over trailing 30- and 60-day windows.
Historical price windows end at the close of the preceding trading day.
The same feature construction is applied to the FNSPID historical data and the FMP test data.

\paragraph{Financial news.}
The historical FNSPID news corpus contains 26,305 unique asset--article records for the selected stock universe.
News for the later test period is obtained from FMP.
For each trading date, the news cutoff is set to 20:00 in the \texttt{America/New\_York} timezone on the preceding day.
We retain at most the two most recent asset-specific records published within the seven days preceding this cutoff.
An asset with no eligible record receives an empty news list.
The timestamp-filtered offline records are fixed for each decision date and shared across methods.

\paragraph{Account context and information availability.}
The trading input also includes the current account state: available cash, existing holdings, net asset value, gross exposure, and net exposure.
The current decision date and the next settlement date are provided, but future settlement prices are excluded.
All decisions, memory retrieval, and financial tool queries are restricted to information available at the corresponding decision cutoff.
Outcome-dependent experience becomes available only after its supporting market outcomes have been observed.

\subsection{Unified Portfolio Construction} 
\label{app:execution}

All active methods use the same asset universe, daily execution frequency, portfolio constraints, and transaction-cost settings.
Each method retains its internal architecture and decision process.
Method-specific adapters translate its native outputs into executable adjustments to the inherited account.
Trading actions and target portfolio weights are interpreted relative to the current holdings.

\paragraph{Return forecasts.}
For methods that predict asset returns, we apply the following return-to-position conversion to each asset.
A predicted return of at least $0.2\%$ produces a long decision, a predicted return of at most $-0.2\%$ produces a short decision, and a prediction between these thresholds produces a flat decision.
For a non-flat decision, the absolute predicted return determines the position-size level $q_{t,j}$ according to Table~\ref{tab:forecast_position_size}, where $t$ indexes the trading step and $j$ indexes the asset.
Flat decisions receive a position-size level of zero.
These conversions use only model predictions and never access realized future returns.

\begin{table}[t]
    \centering
    \caption{Mapping absolute predicted returns to position-size levels.}
    \label{tab:forecast_position_size}
    \begin{tabular}{lc}
        \hline
        Absolute predicted return & Position-size level $q_{t,j}$ \\
        \hline
        $0.2\% \leq |\hat{r}_{t,j}| \leq 0.6\%$ & 1 \\
        $0.6\% < |\hat{r}_{t,j}| \leq 1.2\%$ & 2 \\
        $1.2\% < |\hat{r}_{t,j}| \leq 1.8\%$ & 3 \\
        $1.8\% < |\hat{r}_{t,j}| \leq 2.4\%$ & 4 \\
        $|\hat{r}_{t,j}| > 2.4\%$ & 5 \\
        \hline
    \end{tabular}
\end{table}

\paragraph{Reinforcement-learning actions.}
For an RL policy, the signed action is clipped to $[-1,1]$.
Its sign determines the trading direction, and its absolute magnitude is multiplied by five to obtain a position-size level in $[0,5]$.
A zero action produces a flat position.

\stitle{Direct trading decisions.}
ACE-Trader outputs \texttt{BUY}, \texttt{SELL}, or \texttt{HOLD} decisions.
For assets to be adjusted, \texttt{weight\_pct} specifies the signed target position value as a percentage of the account's net asset value after transaction costs, rather than the percentage to buy or sell.
Positive and negative target weights denote long and short positions, respectively.
These continuous target weights are not discretized into the five position-size levels.

A \texttt{BUY} can open or increase a long position or cover a short position.
A \texttt{SELL} can reduce a long position or open or increase a short position.
The executed quantity is determined by the difference between the target holding and the current holding.
Position retention under \texttt{HOLD} or omission follows the rules below.

For baseline outputs expressed as a target-position direction $s_{t,j}$ and a position-size value $q_{t,j}$, we retain the conversion $w_{t,j}=0.2s_{t,j}q_{t,j}/5$.
Here, $s_{t,j}$ denotes a long, short, or zero target position, rather than a \texttt{BUY}, \texttt{SELL}, or \texttt{HOLD} action.
This conversion is not applied to ACE-Trader's directly specified target weights.

\stitle{Portfolio constraints and position retention.}
Both long and short positions are permitted.
The resulting portfolio, including both adjusted and retained positions, is subject to a gross-exposure limit of 100\% of net asset value.
A \texttt{HOLD} decision or omission of an asset preserves its existing share quantity without trading.
Target weights are applied only to positions explicitly designated for adjustment.

An explicit zero target for a position to be adjusted closes that position.
Neither \texttt{HOLD} nor omission is interpreted as a zero target.
Explicit zero targets produced by a baseline's output adapter remain zero-exposure instructions.

Unknown assets are discarded without affecting existing holdings.
If an output cannot be parsed or contains no valid decision, no trades are executed and the current holdings are retained.
Unallocated capital remains in cash.

\subsection{Trading and Evaluation Protocol}\label{app:protocol}

\stitle{Continuous-account execution.}
Each evaluation trajectory starts without active positions, and its net asset value is normalized to one for reporting.
At each trading step, the trading plan is applied to the inherited account.
For ACE-Trader, only the final memory-assisted plan is executed in the continuous trading account.
The paired shadow executions used for selector supervision do not alter this account, and the memory-free reference does not maintain a separate account across trading steps.

For each position explicitly designated for adjustment, the target position value is obtained from its target weight and the account's net asset value after transaction costs.
Dividing this value by the execution price gives the target share quantity.
Only the difference between the target and current share quantities is traded.
Positions assigned \texttt{HOLD} or omitted from the plan retain their current share quantities and are not rebalanced to their previous portfolio weights.

\paragraph{Transaction costs.}
A one-way transaction-cost rate of 5 basis points is applied to traded notional.
Costs are charged when positions are opened, adjusted, closed, or reversed.
For a retained asset, costs apply only to the executed adjustment.
A position reversal counts both the closing and opening amounts.
No additional full-liquidation cost is imposed at the end of each trading day.
Borrowing fees, additional slippage, and cash interest are not modeled, and unallocated cash earns zero return.
The same cost assumptions apply to every active method.

\paragraph{Chronological evaluation.}
Task-specific offline training uses only the historical training period, with no separate validation set and no method-specific hyperparameter search on the test period.
Baseline configurations follow Appendix~\ref{app:baseline_configurations}.
The test period is processed in chronological order.

For ACE-Trader, the trading LLMs and semantic encoder remain frozen.
New experience is stored only after the associated outcomes become observable, and selector updates use only feedback from completed settlement intervals.
These updates affect subsequent decisions and follow the online learning procedure described in the method section.

\paragraph{Passive benchmarks.}
We additionally report Market Buy-and-Hold, which tracks the Nasdaq Composite over the test period, and Equal-Weight Buy-and-Hold, which allocates capital equally across the 33 stocks at the beginning of the period and holds the resulting portfolio without active rebalancing.
These passive strategies bypass the active decision-conversion pipeline and serve as reference trajectories.

\subsection{Evaluation Metrics}
\label{app:metrics}

We report cumulative return, annualized return, Sharpe ratio, maximum drawdown, annualized volatility, hit rate, and average daily turnover.
Let $T$ denote the number of daily return observations in the continuous account trajectory, and let $r_t^{\mathrm{net}}$ denote the corresponding daily net portfolio return.
The normalized account value after $t$ completed intervals is
$V_t=\prod_{s=1}^{t}(1+r_s^{\mathrm{net}})$, with $V_0=1$.
All portfolio-return metrics use returns after transaction costs, and annualization uses 252 trading days.

\paragraph{Cumulative return.}
Cumulative return is $V_T-1$, measuring the total change in account value over the test trajectory.

\paragraph{Annualized return.}
Annualized return is $V_T^{252/T}-1$, expressing the compounded portfolio return on an annualized basis.

\paragraph{Sharpe ratio.}
The Sharpe ratio is
$\sqrt{252}\,\overline{r^{\mathrm{net}}}/\operatorname{std}(r^{\mathrm{net}})$,
where $\overline{r^{\mathrm{net}}}$ and
$\operatorname{std}(r^{\mathrm{net}})$ are the mean and standard deviation of daily net portfolio returns.
The risk-free rate is set to zero.

\paragraph{Maximum drawdown.}
Maximum drawdown is the largest percentage decline from a preceding peak in account value:
$\max_{1\leq t\leq T}\left(1-V_t/\max_{0\leq s\leq t}V_s\right)$.
Lower values indicate smaller peak-to-trough losses.

\paragraph{Annualized volatility.}
Annualized volatility is
$\sqrt{252}\,\operatorname{std}(r^{\mathrm{net}})$,
measuring the variability of net portfolio returns.

\stitle{Hit rate.}
Hit rate measures directional accuracy over held asset--interval positions.
Let $\rho_{t,j}$ denote the realized return of asset $j$ over the corresponding execution--settlement interval, and let $w_{t,j}$ denote its signed portfolio weight after execution, including positions retained through \texttt{HOLD} or omission.
We compute
\begin{equation}
    \mathrm{HitRate}
    =
    \frac{
        \sum_{t,j}\mathbf{1}[w_{t,j}\rho_{t,j}>0]
    }{
        \sum_{t,j}\mathbf{1}[|w_{t,j}|>0,\ |\rho_{t,j}|>0]
    }.
\end{equation}
Zero-exposure positions and zero-return intervals are excluded from the denominator.
Retained nonzero positions are included even when no trade is executed.
This metric evaluates the direction of the position held, rather than the direction of the current trading action or the fraction of profitable portfolio days.

\paragraph{Average daily turnover.}
For each trading step, turnover $\tau_t$ is the total absolute traded notional divided by the pre-decision account net asset value.
Average daily turnover is
$T^{-1}\sum_{t=1}^{T}\tau_t$.
Both opening and closing transactions are counted, without the conventional one-half factor.

\section{Further Descriptions of Baselines}
\label{app:baselines}

We provide additional descriptions of the 22 active baselines, organized into the same six categories as in the main text, together with two passive benchmarks.
Each method retains its original architecture and learning objective.
We standardize its terminal interface with the trading environment: direct decision generators produce executable trading decisions, while forecasting models and reinforcement-learning policies are connected through the output-conversion procedure described in Appendix~\ref{app:execution}.
All active methods follow the same temporal information cutoff, asset universe, daily execution frequency, transaction costs, and portfolio constraints.

\subsection{Conventional Forecasting Methods}

These methods predict asset returns from structured market features.
They capture nonlinear feature interactions, temporal dependencies, or cross-asset relationships, and their predictions are converted into portfolio decisions through the common execution interface.

\paragraph{LightGBM.}
LightGBM \citep{ke2017lightgbm} is a histogram-based gradient-boosting framework that grows decision trees leaf-wise.
It models nonlinear interactions among market features to produce asset-level return estimates.

\paragraph{XGBoost.}
XGBoost \citep{chen2016xgboost} constructs an additive ensemble of regression trees, with successive trees correcting the prediction errors of the existing ensemble.
Regularization controls model complexity and reduces overfitting.

\paragraph{LSTM.}
LSTM \citep{hochreiter1997long} encodes historical market sequences using gated recurrent memory cells.
Its temporal representation is used to predict the future return of each asset.

\paragraph{GRU.}
GRU \citep{cho2014learning,chung2014empirical} models temporal dependencies through recurrent update and reset gates.
It provides a compact gated alternative to LSTM for learning patterns in historical market sequences.

\paragraph{MLP.}
MLP \citep{hornik1989multilayer} maps flattened rolling market features to asset-level return predictions.
It serves as a feedforward baseline without an explicit temporal or relational modeling component.

\paragraph{TRA.}
TRA \citep{lin2021learning} maintains multiple return predictors associated with different latent trading patterns.
A temporal routing adaptor assigns market samples to the relevant predictors, while an optimal-transport objective encourages them to capture distinct patterns.

\paragraph{HIST.}
HIST \citep{xu2021hist} models cross-stock dependencies through concept-oriented representations.
It combines information associated with predefined and latent market concepts to generate asset-level return forecasts.

\subsection{Reinforcement-Learning Trading Methods}

The FinRL baselines \citep{liu2021finrl} observe structured market and account information and learn trading policies through sequential interactions with a portfolio environment.

\paragraph{FinRL-DQN.}
FinRL-DQN estimates the expected long-term return of discrete trading actions using a deep Q-network.
The policy selects the action with the highest estimated value at each trading step.

\paragraph{FinRL-A2C.}
FinRL-A2C combines an actor that produces trading actions with a critic that estimates state values.
An advantage signal guides the joint optimization of the policy and value networks.

\paragraph{FinRL-PPO.}
FinRL-PPO optimizes its trading policy using a clipped policy-gradient objective.
The clipping mechanism limits excessively large policy changes during training.

\subsection{Frontier General-Purpose LLMs}

These models directly receive the structured market and account context and generate trading decisions without task-specific fine-tuning.
They use the same direct-decision prompt template, temporal cutoff, asset universe, and output schema, without an additional trading-agent workflow.

\paragraph{GLM 5.3 Flash.}
GLM 5.3 Flash\citep{zeng2026glm} is prompted to analyze the supplied market context and directly generate trading actions and position sizes.
Its outputs are evaluated through the common portfolio interface.
\paragraph{DeepSeek V4 Flash.}
DeepSeek V4 Flash \citep{xu2026deepseek} is evaluated as a separate direct-decision model using the same input template and output protocol as GLM 5.3 Flash.
No task-specific parameter updates are performed.

\paragraph{Qwen3.6-35B-A3B.}
Qwen3.6-35B-A3B \citep{team2026qwen3} processes the complete trading prompt and directly produces structured trading decisions.
The pretrained model is used without task-specific fine-tuning.

\subsection{Finance-Specific LLMs}

These baselines use financial-domain pretraining or adaptation to produce forecasts or trading decisions from financial text or market sequences.
Direct decisions are passed to the common portfolio interface, while forecasting outputs are converted into executable trading actions and position sizes.

\paragraph{FinGPT.}
FinGPT \citep{yang2023fingpt} is a finance-oriented language model adapted to financial text and instruction-following tasks.
It receives the structured trading context, and its generated response is interpreted under the common trading-decision schema.

\paragraph{FinCast.}
FinCast \citep{zhu2025fincast} is a foundation model for financial time-series forecasting.
It produces asset-level forecasts from historical market sequences, which are subsequently converted into portfolio decisions.

\paragraph{Kronos.}
Kronos \citep{shi2026kronos} is a pretrained financial time-series model that captures temporal patterns in market sequences and predicts future price or return dynamics.
Its forecasts are evaluated through the same portfolio-construction and execution procedure as the other forecasting baselines.

\subsection{Single-Agent Trading Systems}

These methods organize market analysis and decision making around one trading agent.
Unlike direct prompting, they retain their agentic workflows and support decision making through tools, memory, reflection, or reinforcement learning.

\paragraph{FinAgent.}
FinAgent \citep{zhang2024finagent} combines multimodal market information with tool-assisted analysis, diversified memory, and reflection.
The agent integrates these intermediate analyses into a final trading recommendation.

\paragraph{FLAG-Trader.}
FLAG-Trader \citep{xiong2025flag} combines an LLM-based trading agent with gradient-based reinforcement learning.
Trading feedback is used to optimize the policy associated with the agent's market representations and decision process.

\subsection{Multi-Agent and Expert-Coordinated Trading Systems}

These methods combine the assessments of specialized agents or financial experts through collaboration, debate, or context-dependent coordination.
We retain their original role assignments and coordination mechanisms while evaluating their final recommendations through the common portfolio interface.

\paragraph{TradingAgents.}
TradingAgents \citep{xiao2024tradingagents} distributes market analysis, strategy discussion, and risk assessment across specialized agents.
Their assessments are coordinated through debate and risk-management stages before a final trading decision is produced.

\paragraph{LLMoE.}
LLMoE \citep{liu2025llmoe} uses an LLM-based router to select or combine predefined financial experts according to the current market context.
The expert outputs are aggregated into a final trading recommendation.

\paragraph{TradExpert.}
TradExpert \citep{ding2024tradexpert} assigns different analytical responsibilities or information sources to specialized LLM experts.
A coordinating component integrates their outputs into a unified trading decision.

\paragraph{MM-DREX.}
MM-DREX \citep{chen2025mmdrex} conditions expert coordination on multimodal market information.
Its dynamic-routing mechanism selects or weights specialized LLM experts before aggregating their recommendations.

\subsection{Passive Benchmarks}

\paragraph{Market Buy-and-Hold.}
This benchmark follows the Nasdaq Composite throughout the test period and provides a reference for broad market performance.

\paragraph{Equal-Weight Buy-and-Hold.}
This benchmark allocates capital equally across all candidate stocks at the beginning of the test period and holds the resulting portfolio without active rebalancing.

\section{Implementation Details}
\label{app:implementation}

\subsection{Experimental Environment and Backbones}

Experiments are conducted on Ubuntu 22.04 with 16 NVIDIA
A800-SXM4-80GB GPUs.
Baseline architectures, optimization settings, and inference
configurations are specified in
Appendix~\ref{app:baseline_configurations}.

The Trader uses Deepseek v4 Flash,
and the Red-Team uses Deepseek v4 Flash.
Candidate memories are encoded using
MiniLM.
The trading LLMs and semantic encoder remain frozen throughout historical warm-up and evaluation.
Only the role-specific selector networks receive parameter updates.

\subsection{Historical Warm-up and Initialization}

Historical warm-up uses data from 2021-01-01 to 2023-12-31, without a
separate validation set.
Testing covers 2026-01-01 to 2026-06-30 and follows the data construction
and chronological evaluation protocol described in Appendies ~\ref{app:protocol} and \ref{app:data}.
No method-specific hyperparameter search is performed on the test period.

At the beginning of historical warm-up, the selector networks are randomly
initialized using the default PyTorch initialization, and all private and
shared memory stores are initially empty.
We then process the 2021--2023 warm-up period once in chronological order.
After each completed settlement, the resulting experience is added to
memory and each eligible selector performs one online update using its
current recency-weighted training buffer.

Evaluation starts from an all-cash account with no active positions,
and net asset value is normalized to one for reporting.
Cash, holdings, and account value are subsequently carried forward
without daily account resets.
During testing, memory construction and selector updates use only
outcomes from completed settlement intervals and affect subsequent
decisions.

\subsection{Memory Retrieval and Selection}\label{app:Selector}

\paragraph{Candidate retrieval.}
Each invocation retrieves candidates from the corresponding
role's private memory and the shared case memory.
The retrieval procedure and candidate limits are 10.

\paragraph{Selector architecture.}
The Trader and Red-Team maintain separate selectors, each consisting of
$K=3$ feedforward networks.
The phase-specific retrieval query and candidate memory are encoded into
384-dimensional embeddings using the frozen encoder $E$.
We concatenate the two embeddings and their element-wise product to obtain
a 1,152-dimensional vector.
A linear layer maps this vector to 16 dimensions, followed by a Tanh
activation, producing the semantic features $\mathbf{e}_{t,m}^{u}$
used in Eq.~\ref{eq:memory_representation}.

We then concatenate this 16-dimensional semantic feature with the
corresponding market, portfolio, and recorded-outcome features to form the
complete selector input.
The resulting input dimension is 42 for the Trader and 43 for the Red-Team.
Each complete feature vector is then fed into every network of the
corresponding selector ensemble.

Each Trader network has dimensions
$42\!\rightarrow\!32\!\rightarrow\!16\!\rightarrow\!1$,
whereas each Red-Team network has dimensions
$43\!\rightarrow\!32\!\rightarrow\!16\!\rightarrow\!1$.
The hidden layers use LeakyReLU activations with a negative slope of $0.01$.
The final layer outputs a logit during training, and a sigmoid is applied
at inference time to obtain the predicted probability.

\paragraph{Selection rule.}
Candidates are scored by the ensemble mean prediction plus
the disagreement bonus defined in Section~4.2.
The disagreement-bonus coefficient is
$\alpha=\text{0.8}$.

For every completed selection, the selected candidate and
its selection-time feature vector are saved for subsequent
feedback.

\paragraph{Memory retention and consolidation.}
The Trader and Red-Team maintain separate short-term experience stores and long-term lessons. The short-term memory of each role is capped at 24 experience records. Long-term consolidation is triggered after every 10 newly added short-term experience records. The resulting long-term lessons are stored separately and do not count toward the 24-record short-term memory capacity.

During consolidation, summaries of recent experience are
provided to an LLM to extract recurring lessons with explicit
applicability conditions.
Each lesson retains references to its supporting shared cases
and distinguishes whether a risk occurred from whether the
Trader's revision improved net return.
These references allow either role to revisit the corresponding
interaction without accessing the other role's private memory.

\subsection{Selector Training Settings}
\label{app:selector_training_settings}

We set the feedback tolerance to $\epsilon=0.0005$ (0.05 percentage points), the training window to $W=5$ observed settlement days, and the decay factor to $\gamma=0.8$.
All selector hyperparameters are determined using only the 2021--2023 historical period and fixed before the 2026 test period.
The objective defined in Appendix~\ref{app:selector_update}
is implemented using \texttt{BCEWithLogitsLoss}.
We use Adam with a learning rate of $0.01$, weight decay of
$10^{-4}$, betas $(0.9,0.999)$, and a numerical stability
constant of $10^{-8}$.
Each valid feedback event triggers one gradient update per network.

\subsection{Baseline Configurations}
\label{app:baseline_configurations}

For all open-source baselines, we use their official implementations or faithful interface wrappers, retaining the released architectures, optimizers, learning-rate schedules, and hyperparameters.
Only dataset-dependent input and output dimensions are adapted to the common market features, asset universe, and trading-decision space.
All methods are evaluated through the same portfolio postprocessor and simulator, without method-specific hyperparameter search on the test period.
The configurations are organized into the same six categories as in the main text.

\paragraph{Conventional forecasting methods.}
For LightGBM \citep{ke2017lightgbm} and XGBoost \citep{chen2016xgboost}, we retain their official baseline hyperparameters without task-specific tuning.
LSTM \citep{hochreiter1997long} and GRU \citep{cho2014learning,chung2014empirical} use two recurrent layers, a hidden dimension of 64, and no dropout.
Both models are trained for at most 200 epochs with a learning rate of $10^{-3}$ and a batch size of 800.
For MLP \citep{hornik1989multilayer}, we follow the official Qlib configuration, using a learning rate of $2\times10^{-3}$, a batch size of 4,096, and at most 8,000 optimization steps.
TRA \citep{lin2021learning} employs a two-layer LSTM backbone with a hidden dimension of 64 and three latent trading states.
Its routing adaptor is a one-layer LSTM with a hidden dimension of 32.
The sequence length and batch size are set to 60 and 1,024, respectively.
HIST \citep{xu2021hist} uses a two-layer LSTM backbone with a hidden dimension of 64, mean squared error as the prediction loss, and a learning rate of $10^{-4}$.

\paragraph{Reinforcement-learning trading methods.}
FinRL-DQN, FinRL-A2C, and FinRL-PPO \citep{liu2021finrl} use the official FinRL implementations.
FinRL-DQN retains the default DQN configuration.
FinRL-A2C uses 5 rollout steps, a learning rate of $7\times10^{-4}$, and an entropy coefficient of 0.01.
FinRL-PPO uses 2,048 rollout steps, a batch size of 64, a learning rate of $2.5\times10^{-4}$, and an entropy coefficient of 0.01.

\paragraph{Frontier general-purpose LLMs.}
For GLM 5.3 Flash\citep{zeng2026glm}, DeepSeek V4 Flash \citep{xu2026deepseek}, and Qwen3.6-35B-A3B \citep{team2026qwen3}, we use the official identifiers
\texttt{glm-5.3-flash},
\texttt{deepseek-v4-flash}, and
\texttt{Qwen/Qwen3.6-35B-A3B}, respectively.
All three models use their default reasoning configurations without task-specific fine-tuning.
They receive the same structured trading prompt and produce decisions in the same output format.

\paragraph{Finance-specific LLMs.}
FinGPT \citep{yang2023fingpt} uses the officially released Llama-2-7B LoRA financial-forecaster checkpoint.
FinCast \citep{zhu2025fincast} uses its released pretrained checkpoint under the recommended zero-shot forecasting setting.
Kronos \citep{shi2026kronos} uses the officially released checkpoint and default inference configuration, with temperature 1.0, Top-$p$ 0.9, and one forecast sample.

\paragraph{Single-agent trading systems.}
For FinAgent \citep{zhang2024finagent}, we preserve the officially released market-intelligence, diversified-memory, reflection, and decision modules.
FLAG-Trader \citep{xiong2025flag} uses SmolLM2-135M-Instruct as its backbone.
Its learning rate, rollout length, discount factor, and PPO clipping coefficient are set to $5\times10^{-4}$, 40, 0.95, and 0.2, respectively.
The model is trained for 13,860 steps.

\paragraph{Multi-agent and expert-coordinated trading systems.}
TradingAgents \citep{xiao2024tradingagents} retains its original analyst roles, debate procedure, risk-management module, and default model assignments.
LLMoE \citep{liu2025llmoe} uses a five-day lookback window, a Llama-3.2 router, and two sentiment-oriented experts.
Each expert is a feedforward network with hidden dimensions of 128, 64, and 32, with dropout rates of 0.3 and 0.2 after the second and third layers, respectively.
TradExpert \citep{ding2024tradexpert} employs four specialized experts and one general expert based on LLaMA-2-7B.
Its LoRA rank, scaling factor, and dropout are set to 8, 16, and 0.1, respectively.
Training runs for 30 epochs with a learning rate of $10^{-4}$, a batch size of 4, and 8 gradient-accumulation steps.
MM-DREX \citep{chen2025mmdrex} uses Qwen2.5-VL-72B-Instruct with four expert heads and a 100-day lookback window.
Its LoRA rank, scaling factor, dropout, and learning rate are set to 16, 32, 0.05, and $3\times10^{-5}$, respectively.
The discount factor and policy clipping coefficient are 0.99 and 0.15.

\subsection{Ablation Settings}
\label{app:ablation_settings}

All variants retain the same Commit--Challenge--Revise workflow.
When TS or RS is disabled, we retain the candidate retrieval procedure and supply its top-ranked candidate without learned re-ranking, keeping the one-memory budget unchanged.
Disabling TS applies to both Commit and Revise.
When CE is disabled, each role continues to update and consolidate its private memories from its own outputs and observed outcomes.
Shared interaction records are retained for evaluation but are neither retrieved nor used to construct memories linking the two roles.
The enabled selectors follow the same supervision and update procedure as the complete model.

\end{document}